\documentclass[conference]{IEEEtran}
\IEEEoverridecommandlockouts
\usepackage{cite}
\usepackage{booktabs}
\usepackage{amsmath,amssymb,amsfonts}
\usepackage{algorithmic}
\usepackage{graphicx}
\usepackage{textcomp}
\usepackage{xcolor}
\usepackage{url}
\usepackage{hyperref}
\usepackage{float}
\hypersetup{
    colorlinks=true,
    urlcolor=blue,
}
\def\BibTeX{{\rm B\kern-.05em{\sc i\kern-.025em b}\kern-.08em
    T\kern-.1667em\lower.7ex\hbox{E}\kern-.125emX}}
\usepackage[utf8]{inputenc}
\usepackage{newunicodechar}
\newunicodechar{ }{\,}
\begin{document}

\title{\textbf{FrugalSOT - Frugal Search Over The Models}}
\author{\IEEEauthorblockN{Pradheep P}
\IEEEauthorblockA{\textit{Computer Science and Engineering} \\
\textit{Vellore Institute of Technology}\\
Vellore, India \\
pradheep.p2022@vitstudent.ac.in}
\and
\IEEEauthorblockN{Yuvanesh S}
\IEEEauthorblockA{\textit{Electronics and Communication Engineering} \\
\textit{Vellore Institute of Technology}\\
Vellore, India \\
yuvanesh.s2022@vitstudent.ac.in}
\and
\IEEEauthorblockN{Harish KB}
\IEEEauthorblockA{\textit{Computer Science and Engineering} \\
\textit{Vellore Institute of Technology}\\
Vellore, India \\
harish.kb2022@vitstudent.ac.in}
\and
\IEEEauthorblockN{Keerthan Saai Reddy S}
\IEEEauthorblockA{\textit{Computer Science and Engineering} \\
\textit{Vellore Institute of Technology}\\
Vellore, India \\
keerthansaai.reddys2022@vitstudent.ac.in}
\and
\IEEEauthorblockN{Joshva Devadas T}
\IEEEauthorblockA{\textit{Department of Analytics} \\
\textit{Vellore Institute of Technology}\\
Vellore, India \\
joshvadevadas.t@vit.ac.in}
\and
\IEEEauthorblockN{Naveenkumar J}
\IEEEauthorblockA{\textit{Department of Computational Intelligence} \\
\textit{Vellore Institute of Technology}\\
Vellore, India \\
naveenkumar.jk@vit.ac.in}
\and
\IEEEauthorblockN{Hemalatha K}
\IEEEauthorblockA{\textit{Department of Sensor and Biomedical Technology} \\
\textit{Vellore Institute of Technology}\\
Vellore, India \\
k.hemalatha@vit.ac.in}
}
\maketitle

\begin{abstract}
In on-device NLP tasks, limited resources of embedded hardware, such as the Raspberry Pi 5, require efficient inference strategies. This paper introduces FrugalSOT (Frugal Search Over The Models), a resource-aware model selection architecture for on-device NLP inference. FrugalSOT estimates each request’s complexity by extracting features such as prompt length, named entity density, and syntactic complexity. The request is first made to the least complex model that is likely to pass a relevance threshold. If the output of that model falls short of the threshold, the request is made to a more complex model. It is important to note that the relevance threshold undergoes continuous updates in the background. using past validation outcomes in an adaptation process using a low-pass filtering mechanism, thus imparting adaptation to changing input patterns. Experimental results achieved on a Raspberry Pi 5 show that FrugalSOT reduces average inference time and overall computational resource use to a significant extent compared to a single-model baseline approach, without compromising output relevance to the same extent as the most sophisticated model. These results confirm that adaptive model selection can enable efficient, high-quality natural language processing inference on limited devices.
\end{abstract}


\section{\textbf{Introduction}}Large Language Models (LLMs) have transformed natural language processing (NLP) by driving AI assistants, contextual learning, and document understanding.  Outside NLP, their adaptability is equally applicable in the fields of healthcare (diagnosis, drug discovery), finance (detection of fraud, automated reporting), education (tutoring on a personal level, content creation), and software development (code generation, debugging). But their use of cloud infrastructure for computationally intensive workloads introduces latency, privacy concerns, and accessibility issues, especially for real-time or sensitive use cases such as medical data processing or confidential enterprise workflows. These restrictions strict adoption in resource-scarce settings, such as edge devices or networks with unreliable connectivity, where local inference efficiency matters most.

To solve this, we introduce FrugalSOT (Frugal Search Over The Models), a compact platform that dynamically chooses the best LLMs per request complexity, keeping computational overhead low while maintaining accuracy. By cascading request queries through small on-device models to large models only when necessary, FrugalSOT supports efficient edge deployment referenced on a Raspberry Pi 5, removing cloud reliance. This strategy minimizes latency, improves privacy for use cases, such as using LLMs in healthcare and finance, and extends the reach of LLM capabilities to low-resource settings. Our system employs bridges between various high-performance and real-world, via deployment, unlocking LLM potential across industries without excessive infrastructure demands.

\section{\textbf{Literature Review}}

\subsection{\textbf{Evolution of Large Language Models}}

Building on this foundation, the scale and capability of language models grew rapidly. In 2018–2019, models such as OpenAI’s GPT-2 (1.5B parameters) demonstrated that simply increasing size of the network and training data could drastically improve the quality of text generation. In parallel, BERT (110M parameters) improved many understanding tasks (e.g., GLUE, SQuAD) via deep bidirectional contextual embeddings.The trend was followed by even bigger models: GPT-3 (175B parameters) demonstrated that extremely large transformer models have exceptional performance without task-specific fine-tuning. GPT-3 was fine-tuned once on a huge corpus and was able to translate, answer questions, do arithmetic, and more by conditioning on a few examples given in the prompt. Similarly, Google’s Pathways Language Model (PaLM), with 540B parameters, further confirmed the benefits of scale. PaLM was trained on trillions of tokens and achieved state-of-the-art few-shot results on a wide range of language benchmarks. More recent models have pushed these ideas further. Meta AI’s LLaMA family (2023) includes open-foundation transformer models ranging from 7B to 65B parameters. Remarkably, LLaMA-13B was shown to match or exceed GPT-3 (175B) on many evaluation tasks, despite being an order of magnitude smaller. Other efforts (e.g., BLOOM, GPT-4, etc.) keep investigating scaling and better architectures. Overall, scaling principles imply that larger models having more data are optimal, but newer studies (e.g., Chinchilla models) also look into balancing model size and data for efficiency. These standards identify that modern LLMs are more scalable and capable of processing language, setting new benchmarks in tasks.

\subsection{\textbf{Challenges in Deploying LLMs}}
1. Compute and memory constraints: LLMs have billions of parameters and take gigabytes of memory, which is much more than what is available on most embedded hardware. For example, a 7B parameter LLaMA-2 model takes more than 8 GB of RAM even in half-precision (FP16). The self-attention mechanism of transformer has quadratic complexity (O(n²)) in
input sequence length that induces serious computational and
throughput bottlenecks on low-power CPUs/GPUs. In reality,
modest edge boards (for example, a 4 GB Raspberry Pi 4) can only support
extremely small or highly quantized LLMs.

2. Latency, bandwidth, and privacy: Offloading LLM inference to the cloud introduces network delay and heavy bandwidth use. Cloud APIs for LLMs are noted to suffer from “long response times” and incur significant data-transfer costs, and they require constant connectivity. Such delays often exceed acceptable response times in real-time applications. Additionally, usage of remote servers subjects sensitive information to privacy threats that render pure cloud systems unsuitable for offline or privacy-sensitive environments.

3. Model compression trade-offs: For edge deployment, a smaller model would consume fewer resources; quantization and pruning are the techniques chosen. For instance, weights to 8-bit conversion conserves memory but loses precision (numerical precision loss). Pruning accelerates inference with the loss of partial parameters, whereas unconsidered pruning in Transformers can severely affect performance due to the deletion of some crucial weights. Balancing compression against accuracy is thus a critical deployment challenge.

4. Fine-tuning overhead: Adapting LLMs to specialized tasks on-device is extremely resource-intensive. For instance, Fine-tuning a model with 6.7 billion parameters typically demands around 70 GB of Memory, far beyond the few gigabytes available on most mobile devices. Even parameter-efficient methods (like LoRA) reduce memory by only ~30\%, since gradients must still pass through the full network. Moreover, the backward (training) pass is far more compute-heavy than inference, and most edge accelerators (mobile NPUs/GPUs) lack support for back-propagation. These factors make on-device LLM adaptation slow or infeasible without offloading parts of the computation.

5. Hardware heterogeneity: Edge devices range from ARM smartphones and IoT boards to small GPUs and NPUs, each with different compute cores and memory bandwidth. In practice, inference uses specialized runtimes (e.g. llama.cpp or MLC-LLM on ARM, vLLM on GPUs). This diversity requires tailored software optimizations and hardware–software co-design for each platform. Energy and thermal limits of embedded platforms further constrain model size and concurrency, imposing additional trade-offs in on-device LLM deployment.

\subsection{\textbf{Architectures for Resource-Constrained Environments}}
In bridging the gap in using LLMs in low-resource settings, developers have tried lightweight architecture with an emphasis on performance. Architectures like ALBERT (A Lite BERT), MiniLM, and ELECTRA drive parameter sharing and efficient pretraining methods to shrink model sizes and computation costs significantly. Recent architectures like Mistral and LLaMA 2 Chat utilize low-rank adaptation (LoRA) innovations and sparse attention patterns to result in high performance with less resource usage. In addition, techniques like knowledge distillation enable information transfer from larger models to smaller, more efficient
ones. These have made it possible to implement AI solutions
on edge devices. But trade-offs between reduced complexity
and the ability for executing diverse and complicated tasks require
more investigation to be applicable on a large scale.

\subsection{\textbf{Privacy and Decentralization in AI Systems}}
Decentralized AI systems are made necessary by two factors: privacy and real-time computing. Decentralized AI systems offload computation to edge devices and limit data transfer to the cloud. This reduces exposure of data to breaches and makes it easier to comply with privacy laws. Research determines the potential of edge computing for high-risk domains, such as healthcare and autonomous systems. The implementation of advanced models like LLMs on-premises is, however, subject to overcoming resource and scalability limitations, which are research active.

\section{\textbf{Methodology}}

\subsection{\textbf{User Input and System Setup}}
The process in FrugalSOT starts with a user initiating a prompt either via a graphical interface or by executing the following command in the terminal on Ubuntu 22.04 installed on a Raspberry Pi 5:

\begin{quote}
\ttfamily
frugalsot run 'PROMPT'
\end{quote}
This is what should mimic edge computing scenarios in the wild, where there is limited resource and cloud connectivity may be absent or intermittent. The user input is caught and initiates a sequence of Python and shell scripts to manage the entire workflow directly on the client device, ensuring all processing remains on-device for enhanced privacy, reduces latency, and is standalone, making it particularly well-suited for high-security applications and offline environments. This initial phase sets the stage for efficient, context-aware AI inference at the edge.

\subsection{\textbf{Prompt Complexity Classification}}
The system analyzes prompt complexity through three dimensions—length, named entity recognition (NER), and syntactic structure—using a weighted scoring model to classify prompts as low, mid, or high complexity for optimal resource allocation. This strategy was selected due to the efficiency of extracting these three features with little computational demands while effectively representing the linguistic complexity associated with model resource use. As such, this will work well in real-time edge computing applications. Fig 1 illustrates the complete workflow of this classification system, showing the preprocessing, complexity analysis, scoring, and final decision stages.
\smallskip

\textbf{1. Length Complexity:} Based on word count, with thresholds empirically determined from analysis of 100 diverse prompts:
$$
\text{LC} = 
\begin{cases} 
\text{Low} & \text{if } L \leq 5 \\
\text{Mid} & \text{if } 6 \leq L \leq 10 \\
\text{High} & \text{if } L > 10
\end{cases}
$$

\textbf{2. NER Complexity:} Based on named entity count, with thresholds set considering typical entity density in natural language:
$$
\text{NERC} = 
\begin{cases} 
\text{Low} & \text{if } C_e = 0 \\
\text{Mid} & \text{if } 1 \leq C_e \leq 3 \\
\text{High} & \text{if } C_e > 3
\end{cases}
$$

\textbf{3. Syntactic Complexity:} Based on structural elements, with sentence length threshold of 12 words corresponding to average English sentence complexity:
$$
S = C_c + C_s + \delta(L_s > 12)
$$
where $C_c$ = conjunction count, $C_s$ = subordinate clause count, $\delta(L_s > 12)$ = 1 if average sentence length exceeds 12 words.

$$
\text{SC} = 
\begin{cases} 
\text{Low} & \text{if } S = 0 \\
\text{Mid} & \text{if } 1 \leq S \leq 2 \\
\text{High} & \text{if } S > 2
\end{cases}
$$

\textbf{4. Weighted Final Score:} Each complexity level converts to numerical values (Low=0, Mid=2, High=4):
$$
C = 1 \cdot \text{LC}_{\text{score}} + 2 \cdot \text{NERC}_{\text{score}} + 3 \cdot \text{SC}_{\text{score}}
$$

\textbf{5. Final Classification:}
$$
\text{Overall Complexity} = 
\begin{cases} 
\text{Low} & \text{if } C \leq 4 \\
\text{Mid} & \text{if } 5 \leq C \leq 8 \\
\text{High} & \text{if } C \geq 9
\end{cases}
$$

This systematic approach ensures accurate complexity assessment with minimal computational overhead.
\begin{figure}
    \centering
    \includegraphics[width=1\linewidth]{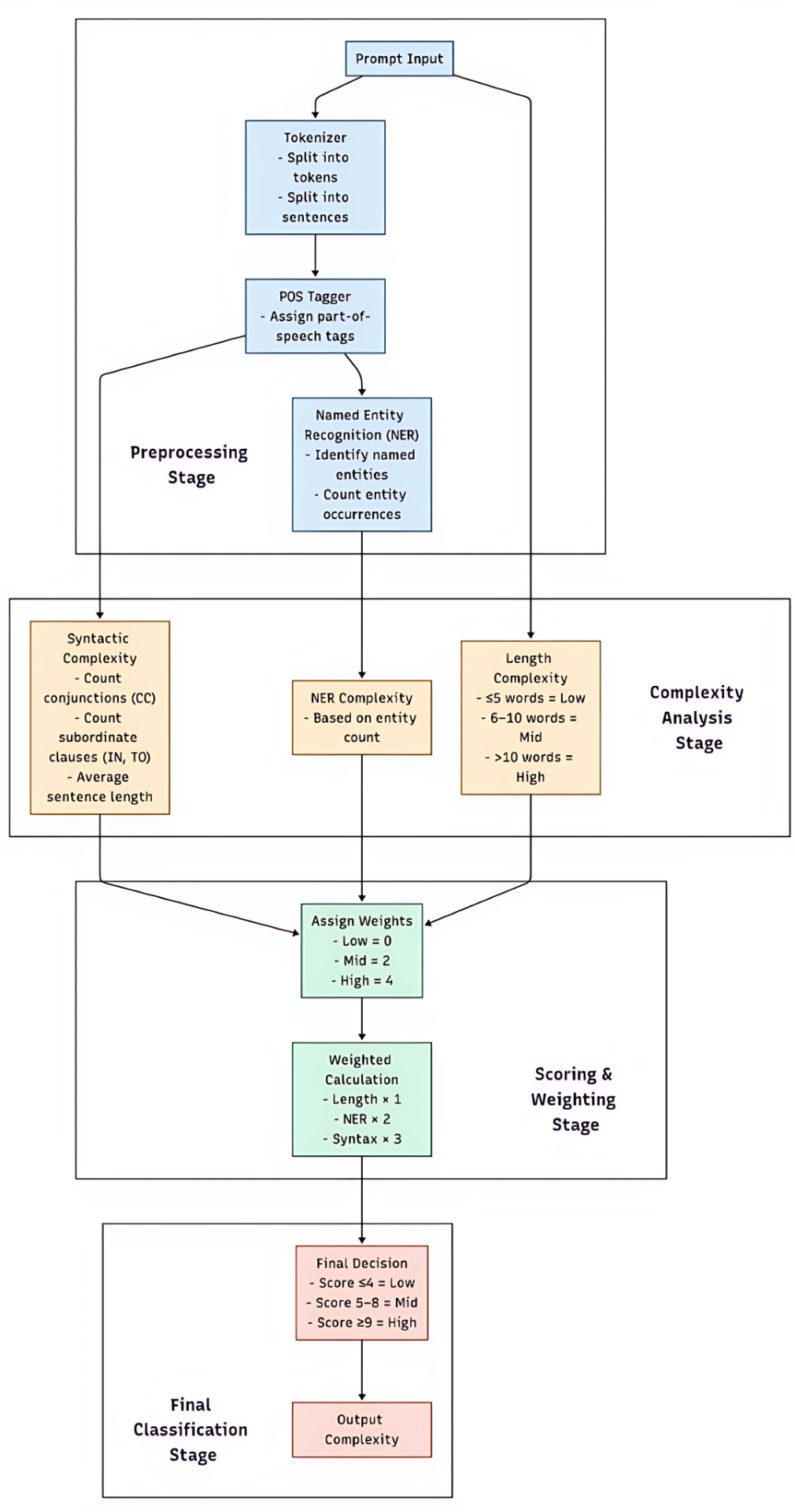}
    \caption{Prompt Complexity Classification}
    \label{fig:prompt_complexity}
\end{figure}

\subsection{\textbf{Dynamic Model Selection}}
The system employs a memory-aware hierarchical model selection strategy, using 8 GB as the threshold based on typical LLM memory requirements and edge device capabilities. For devices with less than 8 GB of memory, resource-optimized models are deployed: TinyLlama (1.1B parameters) for low-complexity queries, TinyDolphin (1.1B parameters) for medium complexity, Gemma2 2b for high complexity, and Phi 2.7b as the fallback model. Devices with more than 8 GB of RAM utilize more capable models: LLaMA2 7b for low complexity, Mistral 7b for medium complexity, LLaMA3.2 3b for high complexity, and LLaMA2 13b as the fallback.

Table \ref{tab:model-selection} summarizes the complete model selection hierarchy across different memory configurations and complexity levels. The Raspberry Pi 5, with 4/8 GB RAM configurations, operates under the lightweight category to ensure stable performance within memory constraints. This cascaded selection minimizes computational overhead by matching model capacity to query complexity, achieving up to 60\% reduction in inference time compared to single-model approaches while maintaining output quality through the intelligent fallback mechanism.

\begin{table}[h!]
\caption{Model Selection Based on Complexity and Available Memory}
\centering
\renewcommand{\arraystretch}{1.4}
\label{tab:model-selection}
\begin{tabular}{lcc}
\toprule
\textbf{Complexity Level} & \textbf{Memory (GB) $<$8} & \textbf{Memory (GB) $>$8} \\
\midrule
Low       & TinyLlama (1.1B)      & LlaMA2 7b      \\
Medium    & TinyDolphin (1.1B)    & Mistral 7b     \\
High      & Gemma2 2b             & LlaMA3.2 3b    \\
Fallback  & Phi 2.7b              & LlaMA2 13b     \\
\bottomrule
\end{tabular}
\end{table}

\subsection{\textbf{Relevance Evaluation and Fallback Mechanism}}
The system assesses output quality through a cosine similarity-based relevancy assessment to the all-MiniLM-L6-v2 model, an embedding model due to its relatively small footprint (22MB, 384 dimensional embeddings), accuracy, and computational output. The relevancy thresholds defined in the threshold analysis were derived from evidence from 250 different prompt-response pairs (100 - low complexity, 100 - medium complexity, and 50 - high complexity), across diverse fields of inquiry with complexity-specific baselines, calculated as averages of the respective similarity scores for each model tier: 0.4441 for low-complexity questions, 0.6537 for medium-complexity questions, and 0.6934 for high-complexity questions. Table \ref{tab:adaptive_thresholds} presents these adaptive complexity thresholds used for real-time system tuning. This extensive dataset ensures statistical robustness and accounts for the inherent difficulty variation across complexity levels, with thresholds undergoing continuous refinement as outlined in Section E, Adaptive Threshold.

\begin{table}[h!]
\caption{Adaptive Complexity Thresholds for Real-Time System Tuning}
\centering
\renewcommand{\arraystretch}{1.3}
\begin{tabular}{lc}
\toprule
\textbf{Complexity Level} & \textbf{Threshold Value} \\
\midrule
Low    & 0.4441 \\
Medium & 0.6537 \\
High   & 0.6934 \\
\bottomrule
\end{tabular}
\label{tab:adaptive_thresholds}
\end{table}

The hierarchical escalation function works like this: when the similarity score of the response from a model drops beneath its threshold for that model's complexity, it will automatically escalate to the next model in the tier hierarchy. For instance, when the scores for TinyLlama from a response or its output was below 0.4441, the system would escalate from TinyLlama to TinyDolphin and forward to Gemma2 2b, and lastly to Phi 2.7b if the assessment of quality could not be satisfied.

This hierarchical fallback is not only used to ensure the accuracy and timeliness of the response. but is also used to achieve maximal computational efficiency. Striking higher up in the hierarchy only when required, the system avoids duplicate computation and conserves power. This is also beneficial in resource-constrained systems, such as devices with less than 8 GB of RAM, such as the Raspberry Pi 5. The fallback model is also a fallback for enabling strong and stable system performance for varying hardware and query complexities.

\subsection{\textbf{Adaptive Threshold}}
The fallback mechanism guarantees that the system escalates to a larger model when the current model ultimately produces an unacceptable result, as defined by the similarity score. We introduce an adaptive threshold to improve over time. The adaptive threshold still relies on the new relevance score and threshold previously defined while using a low pass filter. The low pass filter introduces a threshold that ultimately adapts to user preferences and changing patterns of input and delivers a better fit relative to the altered inputs and expected relevance. \\

The Adaptive Threshold Adaptation follows an exponential moving average:
$$
T_{new} = \alpha \cdot R + (1 - \alpha) \cdot T_{old}
$$
\begin{itemize}
    \item $T_{old}$: The previous value of the threshold before the update.
    \item $T_{new}$: The recalculated threshold value after incorporating the relevance score.
    \item $\alpha$: A factor that determines the influence of the new relevance score on the updated threshold.
    \item $R$: The relevance score generated by the model during its evaluation.
\end{itemize}
We set $\alpha = 0.2$ based on empirical analysis and system stability requirements. This value was chosen because:
\begin{itemize}
    \item Higher $\alpha$ values ($>0.5$) caused threshold oscillations, leading to unstable model switching behavior
    \item Lower $\alpha$ values ($<0.1$) resulted in overly slow adaptation to changing user patterns
    \item $\alpha = 0.2$ provides optimal balance: 20\% weight to new observations and 80\% to historical performance, ensuring system stability while maintaining responsiveness to evolving input complexity
    \item This configuration prevents abrupt threshold changes that could trigger unnecessary model escalations, particularly important for resource-constrained edge devices
\end{itemize}
The low-pass filtering effect of $\alpha = 0.2$ ensures gradual and smooth threshold adjustments, maintaining system stability while continuously improving accuracy. For scenarios where user input evolves over time (e.g., increasing query complexity or changing linguistic patterns), this adaptive mechanism allows the system to remain responsive and relevant without computational overhead.

\subsection{\textbf{Final Output Generation}}
When a response meets the relevance threshold – either from the base or from an escalated model - the system produces the final output to the user. The output consists of the primary response, a confidence score, and a marker indicating the current adaptive threshold status. All outputs and associated metadata are tracked for future tuning and analysis. This end-to-end output generation phase promotes transparency, enables sustained system optimization, and offers users direct, actionable feedback regarding the decision-making process of the system. The outcome is a resilient, resource efficient, and responsive AI inference pipeline that strikes an optimal balance between resource budgets and user-specific high-quality responses.

\section{\textbf{Results}}
{Deploying the adaptive model on the Raspberry Pi 5 brought marked improvements in processing efficiency and response precision. By categorizing incoming queries into three levels of low, mid, and high complexity under a weighted scoring system by prompt length, named entities, and syntactic complexity, FrugalSOT adaptively allocated the least complex suitable model to each query. Light baselines such as tinyllama and tinydolphin performed low- and mid-complexity questions with typical execution durations of approximately 100 and 80 seconds, respectively, to give a rapid response with minimal resource consumption. For high-complexity queries, Gemma2 2b and phi 2.7b were used selectively, with higher execution times (240–280 seconds) but providing more contextual and accurate responses. As illustrated in Fig. 2, critically, FrugalSOT achieved a relevance score of 0.844—comparable to the best-performing individual models—while maintaining an optimized inference time of 200 seconds. This represents a 21.34\% reduction in processing time compared to Phi 2.7b with only a 2.68\% decrease in relevance score, demonstrating the system's ability to balance quality and efficiency through adaptive model selection.}\\

\begin{figure}
    \centering
    \includegraphics[width=1\linewidth]{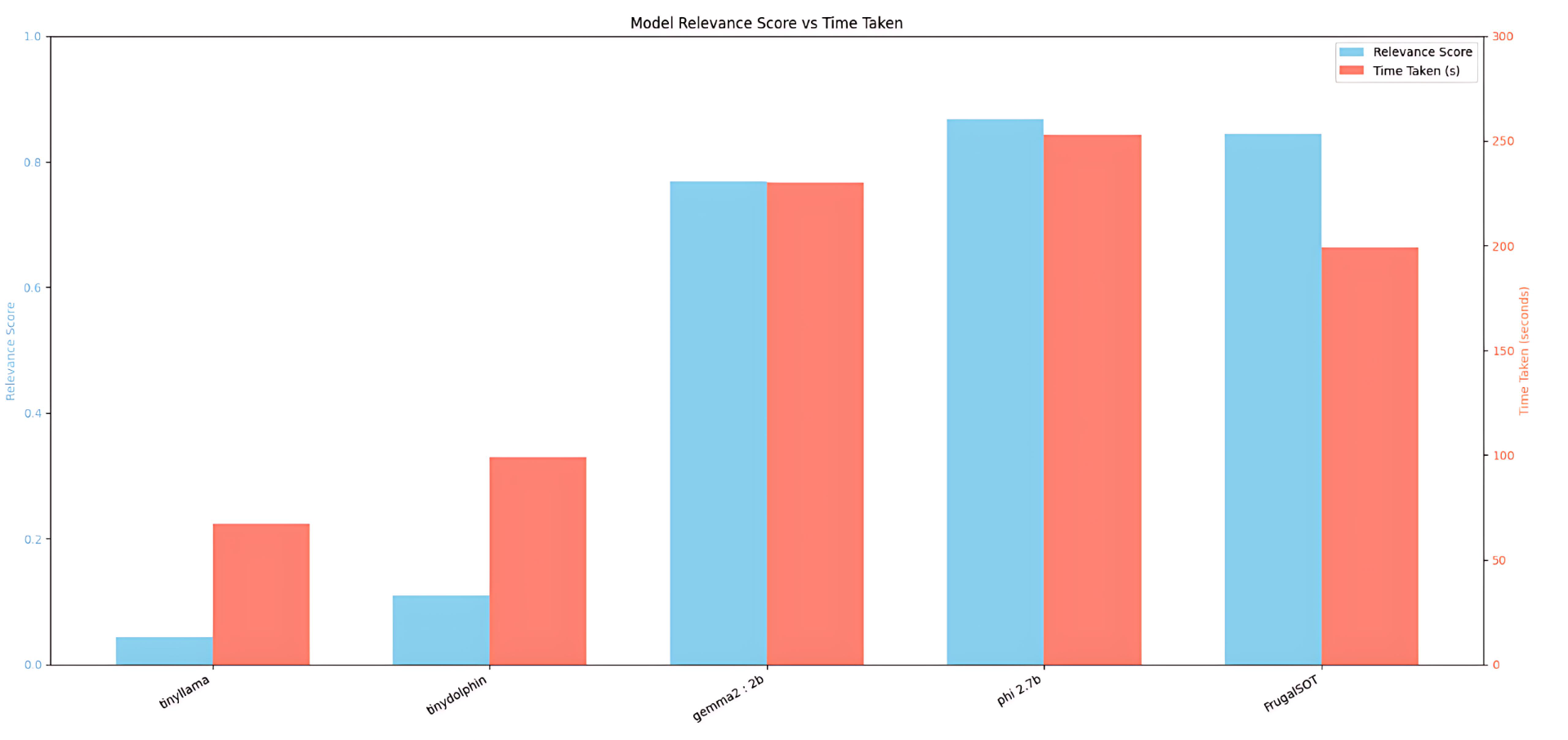}
    \caption{Results}
    \label{fig:results}
\end{figure}

Efficiency analysis also confirmed the ability of FrugalSOT to optimize resource utilization in edge devices. The hierarchical model selection strategy reserved energy-hungry models for resource-constrained queries to conserve energy and computation power. However, Using Ollama for the execution of edge models, however, eliminated the need for cloud connectivity, reducing latency and enhancing data privacy, while relevance-based escalation ensured that more sophisticated models were invoked only if less advanced models were unable to meet acceptable quality levels. In brief, the results verified that FrugalSOT gets a proper balance of speed, power efficiency, and quality of output to be an appropriate candidate for AI inference in low-resource environments such as IoT, autonomous cars, and real-time edge computing applications

\section{\textbf{Conclusion}}
{
FrugalSOT is an efficient, scalable, and adaptable AI inference framework for edge devices that addresses the growing needs of real-time and resource-constrained systems. With dynamic model choice, robust fallback option and adaptive thresholding, FrugalSOT maximizes processing resources, minimizes energy consumption, and delivers relevant high-quality outputs. Its module-based architecture facilitates easy inclusion of upcoming light-weight models, future-proofing the system against changing AI workloads. Together, these combined enable FrugalSOT to provide actionable intelligence across a wide range of applications including intelligent surveillance, autonomous IoT, and environmental monitoring, making it a revolutionary solution for next-generation edge computing installations. Future work will investigate optimization approaches that include model tiling and hardware acceleration to enhance both efficiency and adaptability on edge devices.
}

\section*{\textbf{Acknowledgment}}
The authors would like to thank Dr. Naveenkumar J and Dr. Joshva Devadas T for their insightful guidance and constructive feedback throughout the course of this research. Their unwavering guidance, constant encouragement and invaluable support have been instrumental in making this work possible. The authors are also grateful for the continuous mentorship and expertise provided by these distinguished faculty members, whose contributions were essential to the successful completion of this research endeavor.

\renewcommand{\refname}{

\section*{\textbf{Appendix}}
All experimental results, supplementary materials, and the complete source code for FrugalSOT are made publicly available for reproducibility and further research. These resources, including the GitHub repository, can be accessed through our project website:

\begin{quote}
\url{https://frugalsot.vercel.app/}
\end{quote}

The website hosts detailed documentation, usage instructions, and links to the codebase, supporting transparency and open collaboration for the research community. For the most up-to-date information and ongoing updates, please refer to the website.


\textbf{References}}
\begin{thebibliography}{00}
\bibitem{b1} Naveed, Humza, et al. "A Comprehensive Overview of Large Language Models." arXiv preprint arXiv:2307.06435 (2023).

\bibitem{b2} Zhou, Zixuan, et al. "A Survey on Efficient Inference for Large Language Models." arXiv preprint arXiv:2404.14294 (2024).

\bibitem{b3} Wang, Wenhui, et al. "MiniLM: Deep Self-Attention Distillation for Task-Agnostic Compression of Pre-Trained Transformers." Advances in Neural Information Processing Systems 33 (2020): 5776-5788.

\bibitem{b4} Hu, Edward J., et al. "LoRA: Low-Rank Adaptation of Large Language Models." International Conference on Learning Representations (ICLR), 2022.

\bibitem{b5} Dettmers, Tim, et al. "QLoRA: Efficient Finetuning of Quantized LLMs." Advances in Neural Information Processing Systems 36 (2023): 10088-10115.

\bibitem{b6} Hinton, Geoffrey, Oriol Vinyals, and Jeff Dean. "Distilling the Knowledge in a Neural Network." arXiv preprint arXiv:1503.02531 (2015).

\bibitem{b7} Li, Shiyao, et al. "Evaluating Quantized Large Language Models." arXiv preprint arXiv:2402.18158 (2024).

\bibitem{b8} Gholami, Amir, et al. "A Survey of Quantization Methods for Efficient Neural Network Inference." Low-power Computer Vision. Chapman and Hall/CRC, 2022. 291-326.

\bibitem{b9} Kim, Sehoon, et al. "Full Stack Optimization of Transformer Inference: A Survey." arXiv preprint arXiv:2302.14017 (2023).

\bibitem{b10} Liu, Song et al. "An Efficient Tile Size Selection Model Based on Machine Learning." Journal of Parallel and Distributed Computing 121 (2018): 27-41.

\bibitem{b11} Chen, Lingjiao, Matei Zaharia, and James Zou. "FrugalGPT: How to Use Large Language Models While Reducing Cost and Improving Performance." arXiv preprint arXiv:2305.05176 (2023).

\bibitem{b12} "FrugalSOT: Frugal Search Over The Models." FrugalSOT Project Website. https://frugalsot.vercel.app/ (accessed May 20, 2025).

\bibitem{b13} Radford, Alec, et al. "Language Models are Unsupervised Multitask Learners." OpenAI Blog 1.8 (2019): 9.

\bibitem{b14} Devlin, Jacob, et al. "BERT: Pre-training of Deep Bidirectional Transformers for Language Understanding." arXiv preprint arXiv:1810.04805 (2018).

\bibitem{b15} Wang, Alex, et al. "GLUE: A Multi-task Benchmark and Analysis Platform for Natural Language Understanding." arXiv preprint arXiv:1804.07461 (2018).

\bibitem{b16} Rajpurkar, Pranav, et al. "SQuAD: 100,000+ Questions for Machine Comprehension of Text." arXiv preprint arXiv:1606.05250 (2016).

\bibitem{b17} Brown, Tom B., et al. "Language Models are Few-shot Learners." Advances in Neural Information Processing Systems 33 (2020): 1877-1901.

\bibitem{b18} Chowdhery, Aakanksha, et al. "PaLM: Scaling Language Modeling with Pathways." arXiv preprint arXiv:2204.02311 (2022).

\bibitem{b19} Touvron, Hugo, et al. "LLaMA: Open and Efficient Foundation Language Models." arXiv preprint arXiv:2302.13971 (2023).

\bibitem{b20} Scao, Teven Le, et al. "BLOOM: A 176B-parameter Open-access Multilingual Language Model." arXiv preprint arXiv:2211.05100 (2022).

\bibitem{b21} Lan, Zhenzhong, et al. "ALBERT: A Lite BERT for Self-supervised Learning of Language Representations." arXiv preprint arXiv:1909.11942 (2019).

\bibitem{b22} Clark, Kevin, et al. "ELECTRA: Pre-training Text Encoders as Discriminators Rather than Generators." arXiv preprint arXiv:2003.10555 (2020).

\bibitem{b23} Jiang, Albert Q., et al. "Mistral 7B." arXiv preprint arXiv:2310.06825 (2023).

\bibitem{b24} Mangrulkar, Sourab, et al. "PEFT: State-of-the-art Parameter-efficient Fine-tuning Methods." GitHub repository (2022).

\bibitem{b25} Wolf, Thomas, et al. "Transformers: State-of-the-art Natural Language Processing." Proceedings of the 2020 Conference on Empirical Methods in Natural Language Processing: System Demonstrations (2020): 38-45.

\bibitem{b26} Hong-Linh, Truong, Tram Truong-Huu, and Tien-Dung Cao. "Making Distributed Edge Machine Learning for Resource-Constrained Communities and Environments Smarter: Contexts and Challenges." Journal of Renewable Energy, Sustainability and the Environment 40, no. 3 (2022): 176-203.

\bibitem{b27} Yi, Ding, et al. "JMDC: A Joint Model and Data Compression System for Deep Neural Networks Collaborative Computing in Edge-Cloud Networks." Journal of Parallel and Distributed Computing 171 (2023): 24-36.

\bibitem{b28} IEEE Computer Society. "Adaptive Local Thresholding by Verification-Based Multithreshold Probing." IEEE Transactions on Pattern Analysis and Machine Intelligence 25, no. 1 (2003): 131-137.

\bibitem{b29} Zhang, Wei, et al. "A Comprehensive Review of Model Compression Techniques in Machine Learning." Applied Intelligence 54, no. 8 (2024): 5747-5776.

\bibitem{b30} Rahaman, Habibur, et al. "Artificial Intelligence and Edge Computing for Machine Maintenance in Industry 4.0." Applied Intelligence 52, no. 15 (2024): 10748-10782.

\bibitem{b31} Dong, Rongkang, Yuyi Mao, and Jun Zhang. "Resource-Constrained Edge AI with Early Exit Prediction." arXiv preprint arXiv:2206.07269 (2022).

\bibitem{b32} Xu, Hao, et al. "Camel: Energy-Aware LLM Inference on Resource-Constrained Devices." arXiv preprint arXiv:2508.09173 (2025).

\bibitem{b33} Wang, Xubin, and Weijia Jia. "Optimizing Edge AI: A Comprehensive Survey on Data, Model, and System Strategies." arXiv preprint arXiv:2501.03265 (2025).

\bibitem{b34} Jalaian, Brian, et al. "On Accelerating Edge AI: Optimizing Resource-Constrained Environments." arXiv preprint arXiv:2501.15014 (2025).

\bibitem{b35} Hanafy, Walid, et al. "MEL: Multi-level Ensemble Learning for Resource-Constrained Environments." arXiv preprint arXiv:2506.20094 (2025).

\bibitem{b36} Shamatrin, Dmytro. "Adaptive Thresholding for Multi-Label Classification via Global-Local Signal Fusion." arXiv preprint arXiv:2505.03118 (2025).

\bibitem{b37} Zeng, Xinyue, et al. "LENSLLM: Unveiling Fine-Tuning Dynamics for LLM Selection." arXiv preprint arXiv:2505.05375 (2025).

\bibitem{b38} Zhou, Jianpeng, et al. "Adaptive-Solver Framework for Dynamic Strategy Selection in Large Language Model Reasoning." arXiv preprint arXiv:2310.01446 (2023).

\bibitem{b39} Wang, Ruoyu, et al. "DICE: Dynamic In-Context Example Selection in LLM Agents via Efficient Knowledge Transfer." arXiv preprint arXiv:2507.23554 (2025).

\bibitem{b40} Lumer, Elias, et al. "ScaleMCP: Dynamic and Auto-Synchronizing Model Context Protocol Tools for LLM Agents." arXiv preprint arXiv:2505.06416 (2025).

\end{thebibliography}
\end{document}